\documentclass[letterpaper, 10 pt, conference]{ieeeconf}  

\IEEEoverridecommandlockouts                              
                                                          
\usepackage{graphicx}
\usepackage{url}
\title{{\LARGE \bf
Car Type Recognition with Deep Neural Networks}
}
\author{Heikki Huttunen$^{1,2}$ and Fatemeh Shokrollahi Yancheshmeh$^{1}$ and Ke Chen$^{1}$\\
$^{1}$Tampere University of Technology, Finland\\
$^{2}$Visy Oy, Tampere, Finland}

\begin{document}

\thispagestyle{empty}
\pagestyle{empty}

\maketitle
\begin{abstract}
In this paper we study automatic recognition of cars of four types:
Bus, Truck, Van and Small car. For this problem we consider
two data driven frameworks: a deep neural network and a support vector
machine using SIFT features.
The accuracy of the methods is validated with a database 
of  over 6500 images, and the resulting prediction 
accuracy is over 97 \%. This clearly exceeds the accuracies of earlier
studies that use manually engineered feature extraction pipelines.
\end{abstract}


\section{Introduction}

Traffic counting is a central tool for traffic planning and analysis
for intelligent traffic. In many cases the desire is to increase the level
of details by additional categorization of vehicle
types. This allows a more fine-grained analysis and more accurate profiling
of users of the transportation infrastructure, which is necessary for 
assessing the effects of possible modifications in the traffic system.

\begin{figure*}[t]
\centerline{\includegraphics[width=0.4\textwidth]{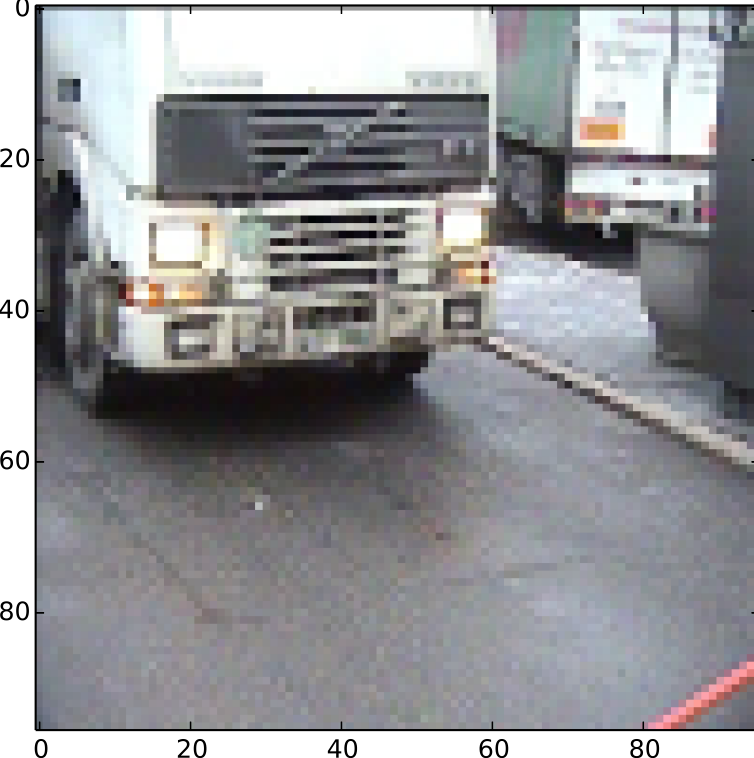}\quad
\includegraphics[width=0.4\textwidth]{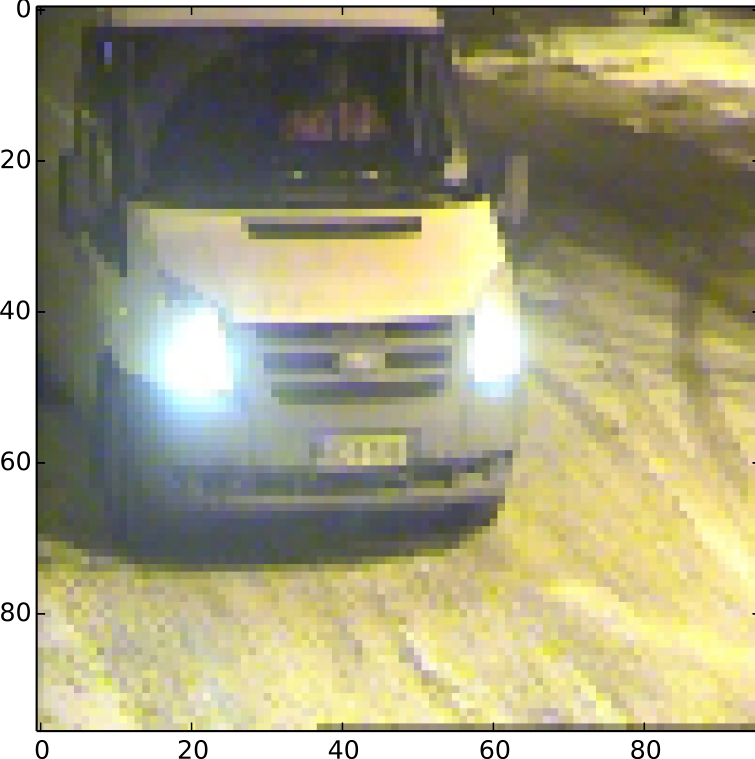}}
\caption{Example images from the two cameras. The images are downscaled to 
$96\times 96$ pixels to illustrate the network input resolution.}
\label{fig:images}
\end{figure*}

There are several production-level techniques for recognition of vehicle types,
including inductive ground loops (see, \emph{e.g.}, 
MAVE-L product line of AVE GmbH\footnote{\url{http://www.ave-web.de/}}) and
laser scanners (see, \emph{e.g.}, traffic counters 
from SICK GmbH\footnote{\url{http://www.sick.de/traffic}}). However, all these
technologies require laborious installation and significant amount of costly
hardware. In this paper we study the use of low cost cameras for
car type classification. The benefits are obvious: Cameras are
ubiquitous and cost-efficient tools for monitoring, and they often
have also other surveillance uses simultaneously. 
However, the reliability of camera based technologies may be vulnerable to
environmental factors, such as poor illumination, dirt or change of
viewing angle, and the developed system should therefore be robust
to such changes. A typical scenario uses camera based recognition
for surveillance, categorizing the entering vehicles at the gate to
an access controlled facility. In this case, the input to the algorithm
would typically consist of frontal view pictures, such as 
those shown in Figure \ref{fig:images}.

Earlier research related to the topic is divided into two main areas:
\emph{car make recognition}~\cite{Car-Make-Hsiao-2014,dlagnekov2005recognizing}
and \emph{car type recognition} that has
received a lot of attention during the recent years,
\cite{kafai2012dynamic,zhang2013vehicle,Lee-Neural-2013}.
From customer (traffic operator) viewpoint, the car type is 
typically more interesting than the car make, which is also our
motivation to study this problem.

Kafai \emph{et al.} \cite{kafai2012dynamic} propose to use a Bayesian network for the
vehicle type classification, with features extracted from 
images of the rear of the car. The resulting feature vector
consists of a collection of geometric parameters of the vehicle;
including simple features such as the vehicle width and height and
more complicated features, such as the distance from the license plate to 
the tail lights. Subsequently, the most significant features from the
pool are fed to a Bayesian network for classification.
The authors report 10-fold cross-validated classification accuracy 
of 95.7\% for a database of 177 vehicles from four categories.

Zhang \emph{et al.} \cite{zhang2013vehicle} propose a framework, where the reconstruction 
error of a vector quantization representation is used for car type
classification. More specifically, a tight bounding box of the
car is first extracted, and a codebook for each class is learned
from the training data. The reconstruction error is used as
a basis for a measure for the class confidence, and
falsely detected object candidates are rejected by thresholding the
classification error with a manually set limit. The authors
report an accuracy of 92 \% and over 95 \% when the rejection
heuristic is used. The database consists of over 2800 images.

In \cite{Lee-Neural-2013}, the authors use a neural network
for car type classification. The input to the network consists of 
a collection of image based features, such as width, height, perimeter
and fractal dimension. The actual classification is done in
two stages using two neural networks, and the authors report
69 \% accuracy on a database with 100 vehicles.

All the above works rely on a more or less elaborate preprocessing
step; at least an accurate vehicle detection and alignment resulting in
a tight bounding box is required. In our work, we wish to avoid all
preprocessing steps, as they may be computationally expensive 
and---more importantly---fragile to errors that may collapse the 
subsequent classification completely. Moreover, we will be considering
an application, where high accuracy is required in various environmental
conditions. In particular, poor illumination, dirt and snow are
typically difficult for preprocessing steps, including detection and
localization.

Our approach compares two data driven frameworks: a deep neural network
and a support vector machine with Scale-Invariant Feature Transform (SIFT) 
image features. In other words,
the proposed methods require no preprocessing except straightforward 
brightness normalization.
The obvious benefits of such an approach are due to the simplicity
of the implementation. 

All the approaches found in the literature concentrate on recognition from still images.
However, a typical access control setup extracts individual frames from a video
stream for recognition. In this paper, we also limit our study to individual
shots from the stream, but deep neural networks can be extended
to process temporal video streams, as well. A straightforward approach is to 
compute the likelihoods for each car type for a number of video
frames. Computing the average (or maximum) of these likelihoods tends to improve 
the accuracy of recognition, since a larger number of frames has a higher probability
of a good frontal view of the vehicle (avoiding \emph{e.g,} cases with car shown only
partially, blocked by another vehicle or changing the lane). Alternatively, the temporal stack of video frames can be 
fed to a deep network directly; see, \emph{e.g.,} Tran \emph{et al.} for categorization
of sports activities \cite{tran2015learning}. 

\begin{table*}[bt]
\caption{Hyperparameters randomized over the iterations. The rightmost column describes the best configuration found within the 50 randomly selected configurations (see text).}
\label{tab:hyperparameters}
\begin{center}
\begin{tabular}{lcc}
\hline
\textbf{Hyperparameter}                                   & \textbf{Range} & \textbf{Selected Value} \\ \hline
\textit{Number of Convolutional Layers}                   & 1 -- 4  &  2 \\                              
\textit{Number of Dense Layers}                           & 0 -- 2  & 2                              \\
\textit{Input Image Size}                                 & \{64, 96, 128, 160\} & 96                \\
\textit{Kernel Size on All Convolutional Layers}          & \{5, 9, 13, 17\} & 5                  \\
\textit{Number of Convolutional Maps} & \{16, 32, 48\}               & 32      \\
\textit{Learning rate}                                    & $10^{-5}$ -- $10^{-1}$                      &0.001643
\end{tabular}
\end{center}
\end{table*}

The remainder of this paper is organized as follows. Section \ref{sec:methods} 
describes the proposed method in detail and in Section 
\ref{sec:experiments}, we report the experimental results and the
image database used for experiments. Finally, Section \ref{sec:conclusions}
will discuss the results and draw conclusions for future work.

\section{Materials and Methods}
\label{sec:methods}

In this Section we will describe two data driven alternatives for car
type recognition. The first one represents a deep architecture and
the second a shallow one. Before describing the architectures, we 
describe the data used later in the experiments.

\subsection{Data}
\label{sec:data}

The database used in this paper
was collected in collaboration with the company Visy Oy\footnote{\url{http://www.visy.fi/}}, whose license plate
recognition based access control system installations have gathered tens of millions
of vehicle images over the years. Each access control checkpoint is equipped with
a digital PAL resolution camera, infrared illumination and a ground loop
for detection of the vehicle in front of the camera.

The database consists of altogether 6555 vehicle images. The database 
was collected at two entrance gates at an access control system installation
in a port in Finland. The pictures were
acquired at interlaced PAL image resolution of $768\times 288$ per field.
The images were extrapolated to the correct aspect ratio and resolution
$768\times 576$. Examples of images from the database are shown in Figure \ref{fig:images},
downscaled to the DNN input resolution of $96 \times 96$.

We experimented with two installation scenarios: In the first case, the
database consists of all images from one checkpoint and one camera. In
this case the background is relatively steady and cars enter from approximately
the same direction and angle. Note, however, that the images are collected over
a long period of time, and there are significant changes in the 
environmental conditions (rain, snow, day/night).
This experiment investigates the performance for
an individual gate installation, which would always be calibrated 
(trained) individually.

The another scenario is more realistic, in that the database consists of
images from two cameras at two different entrance checkpoints. 
Thus, the cars are entering at different angles from different 
directions and the background is not constant. This way we can also
determine whether the methods learn the appearance of the background or the appearance
of the actual vehicle. Namely, it could be possible that the classifier
would learn to recognize small cars by spotting the background, which
is not visible when blocked by a large vehicle.

\begin{figure*}[tb]
\centerline{\includegraphics[width=0.95\textwidth]{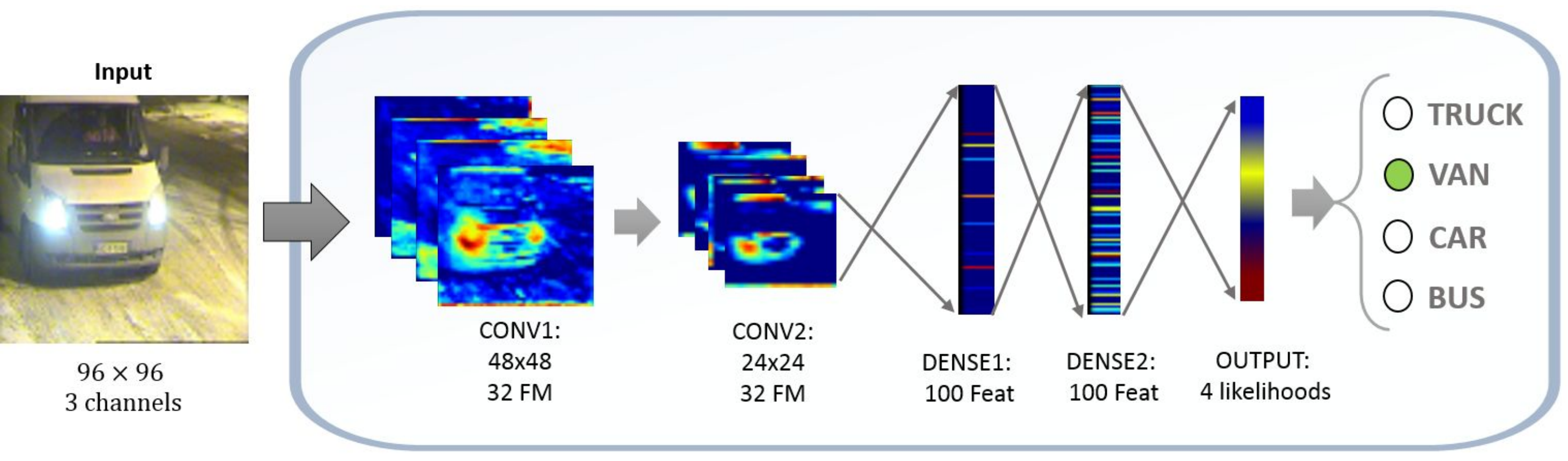}}
\caption{The structure of the proposed network. The network consists of altogether
four hidden layers: two convolutional layers followed by two dense layers and an output layer.}
\label{fig:network}
\end{figure*}

\subsection{Deep Architecture: Deep Neural Networks}
\label{sec:DNN}

This decade has seen a breakthrough in image classification due
to the advances in large neural networks. Several factors have
contributed to their enormous success, including both the explosion of
computational power brought by current Graphics Processing
Units (GPU's), and theoretical advances in neural network
community, that have enabled the training of networks with
very large number of stacked layers (Deep Neural Networks; DNN's). 

\begin{table*}[t]
\caption{Accuracy of classification for the proposed methods. The experiments are done for two cases: A dataset consisting of pictures from a single camera angle, and a second dataset with pictures from two cameras with difference camera angles.}
\centering
\begin{tabular}{lcc}
\hline
 \textbf{Classifier}  & \textbf{Accuracy} & \textbf{\#vehicles} \\
                             \hline
\textit{Deep Neural Network (1 camera)} & \textbf{98.06 \%} &   1500      \\ 
\textit{SIFT + SVM (1 camera)} & 97.35 \% & 1500      \\
\textit{Deep Neural Network (2 cameras)} & \textbf{97.75 \%} & 6555      \\
\textit{SIFT + SVM (2 cameras)} & 96.19 \% & 6555      \\
\hline
\emph{Kafai \emph{et al.} \cite{kafai2012dynamic}} & 95.7 \% & 177 \\
\emph{Zhang \emph{et al.} \cite{zhang2013vehicle}}  &  95 \% & 2800 \\
\emph{de S. Matos \emph{et al.} \cite{Lee-Neural-2013}} & 69 \% & 100\\
\hline
\end{tabular}
\label{tab:results}
\end{table*}

Among the most significant highlights are achievements such as the large scale
image classification record with the ImageNet database \cite{krizhevsky2012imagenet}, the DeepFace face recognition method by Facebook \cite{taigman2014deepface} and the deep network learning
to play computer games by Google \cite{mnih2015human}. Over the years, networks
have grown in depth, ultimately reaching depths of 20-30 layers; see \emph{e.g.},
the 22-layer GoogLeNet \cite{szegedy2014going} reaching the state-of-the art of the 
ImageNet Large-Scale Visual Recognition Challenge 2014 (ILSVRC 2014).
There are several software platforms, where
training and classification of a deep neural network are 
straightforward engineering tasks: These include \emph{Torch7} \cite{collobert2011torch7},
\emph{Keras} \cite{chollet2016} and 
\emph{Caffe} \cite{jia2014caffe}. In our work, we will 
use the latter platform due to its particular efficiency for
\emph{image} data.

The key benefit of adding layers to the network is that this enables the
classifier to learn higher lever structures within the image. For our particular
problem, the network might have the ability to learn the appearance of the 
headlights on the first layer, their relative positions and distance on the second
layer, and their relative location with respect to the car license plate on
the third convolutional layer. In other words, the key to success is that
the network can learn the feature extraction step in an optimal manner and
can avoid the need for manual feature engineering; a critically important
step in most earlier car type recognition approaches
\cite{kafai2012dynamic,zhang2013vehicle,Lee-Neural-2013}.

As a drawback of the deep neural network, there is a need to
define the hyperparameters of the network. The choices of the
number of layers, number of nodes, sizes of convolutional kernels, etc.
all have a crucial importance on the resulting accuracy.
For this problem, we follow the approach of Bergstra \emph{et al.}
\cite{bergstra2012random}, which randomly searches for a good combination
of selected hyperparameters. The authors were able to show that as few as 
8 or 16 iterations with random selection of hyperparameters can outperform 
both manual search and grid search with the same computational budget.

In our case, the hyperparameters randomized in search for the best network
topology are summarized in Table \ref{tab:hyperparameters}. The first two
items define the depth of the network: The structure always consists of
1-4 convolutional layers followed by 0-2 dense layers. The input images
are resized to square shape with dimensions within the range 64-160. The image
size is closely related to the size of the convolutional kernels.
We limit these to be equal in size on all layers  within the range 5-17 pixels
along both axes. The penultimate parameter of Table \ref{tab:hyperparameters}
defines the number of 
convolutional maps (\emph{i.e.,} the number of filters learned at each
layer), and can change between 16-48 maps. Finally, a crucial parameter
to the performance is the learning rate for the stochastic gradient
backpropagation. This parameter is randomly sampled from a geometric distribution
between $10^{-5}$ -- $10^{-1}$.
The rightmost column of Table \ref{tab:hyperparameters} tabulates the selected
hyperparameters in the experiments of Section \ref{sec:experiments}, after
training a network with 50 randomly selected hyperparameter combinations.

The corresponding network topology is illustrated in Figure \ref{fig:network}.
In summary, our car type recognition network consists of five layers;
two convolutional layers followed by two dense layers and an output layer. 
The first convolutional layer maps the three-channel $96\times 96$ 
input into 32 feature maps which are max-pooled to $48\times 48$ resolution.
The second convolutional layer produces another 32 feature maps which 
are then downsampled to $24\times 24$ with max-pooling.
After the convolutional layers, there are two fully connected layers with
100 nodes each. Finally, the output layer maps the 100 features on the
last dense layer into four class likelihoods via a softmax operator.
Between each layer, there is additionally a combination of a 
Dropout regularizer and a Rectified Linear Unit (ReLU) nonlinearity.

\subsection{Shallow Architecture: Support Vector Machine}
\label{sec:svm}

For comparison, we also employ a conventional shallow method: Dense SIFT \cite{vedaldi08vlfeat} and support vector machines (SVM) \cite{cortes1995support}, which is a widely adopted combination in various frameworks for visual recognition such as fine-grained pet classification \cite{parkhi12a} and large-scale image categorization \cite{Chatfield11}. 

We adopt a bag-of-words model \cite{csurka2004visual}, whose visual words are generated densely by extracting SIFT descriptors \cite{lowe1999object} on the grids of images. The hyperparameter setting of dense SIFT follows that of \cite{parkhi12a}, which adopts a stride of 6 pixels and at 4 scales (\emph{i.e.}, the spatial range of bins are 4, 6, 8, and 10 pixels). For incorporating spatial information, we use a spatial pooling method \cite{lazebnik2006beyond}, which divides the whole image region into $1\times 1$ and $2\times 2$ cells. For each cell, SIFT descriptors are quantized into a 1000-clusters vocabulary by K-means clustering. As a result, we have a 5000-dimensional feature vector, whose histogram bins are normalized by $\ell_1$ norm, to represent each image. With the resulting image feature representation and the corresponding car type labels, a multiclass support vector machine using the RBF kernel is applied during training. For an unseen image, the image feature is fed into the trained support vector machines to classify its car type.

\section{Experiments}
\label{sec:experiments}

In this section we study the accuracy of the two classification methods
in Sections \ref{sec:DNN} and \ref{sec:svm}.

\begin{figure}[t]
\centerline{\includegraphics[width=\columnwidth]{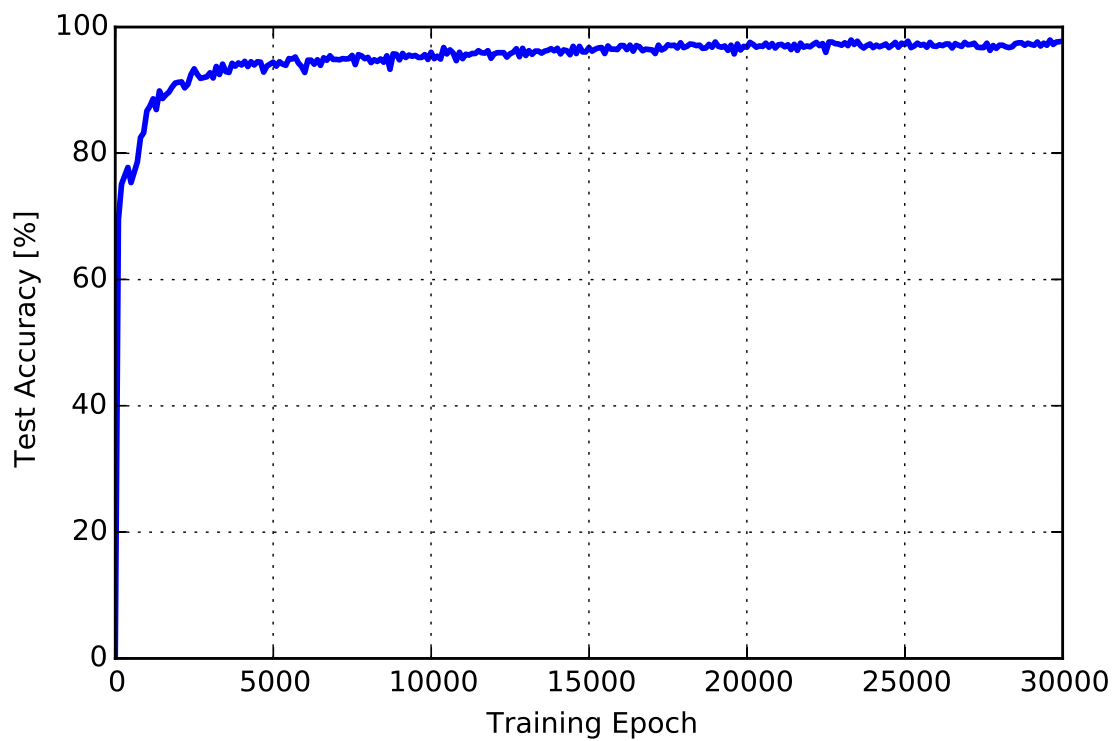}}
\caption{The learning curve of the network.}
\label{fig:accuracy}
\end{figure}

In order to assess the accuracies, the database described in Section \ref{sec:data} 
was split into a training
set (90 \% of all samples) and a test set (10\% of all samples) in 
a stratified manner.

The learning curve of the proposed neural network is shown in Figure~\ref{fig:accuracy},
which shows the classifier accuracy for the test data. The training
is continued for 30,000 epochs without any accuracy related stopping
criterion. The curve clearly shows that the dropout regularization and data
augmentation are effective in avoiding overlearning. Moreover, a good
classification accuracy is reached already at 15,000 iterations.
In total, the training takes approximately 20 minutes on a NVidia
Tesla K40t GPU.

\newcommand{\wid}{0.18\textwidth}

\begin{figure*}[t]
\begin{center}
\includegraphics[width=\wid]{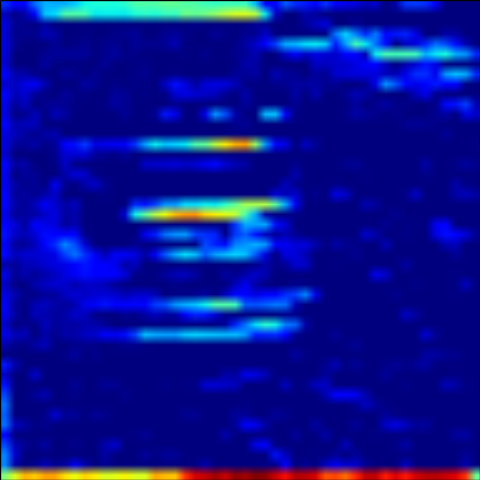}\quad
\includegraphics[width=\wid]{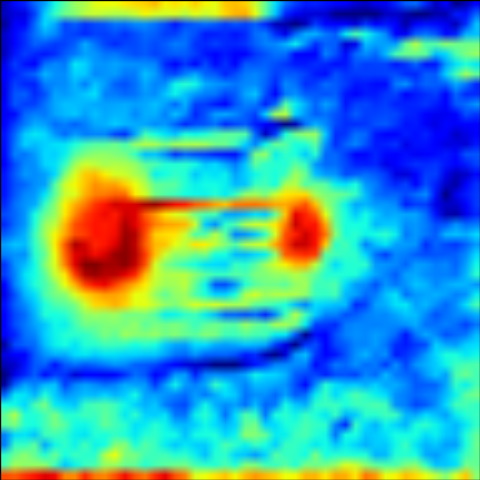}\quad
\includegraphics[width=\wid]{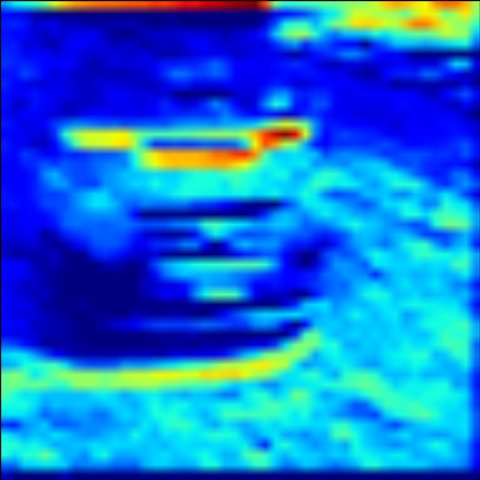}\quad
\includegraphics[width=\wid]{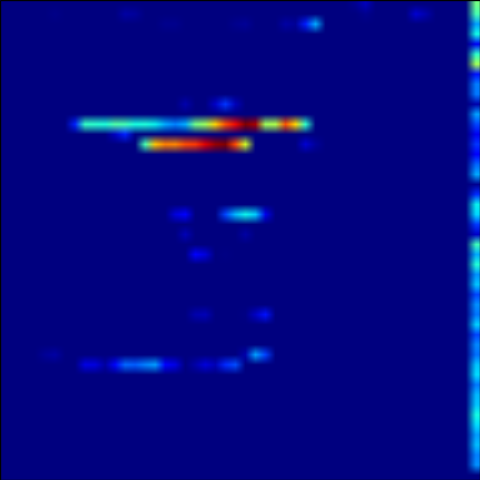}\quad
\includegraphics[width=\wid]{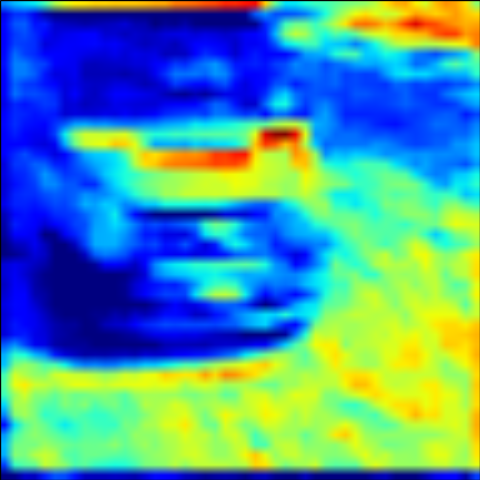}
\\
\smallskip
\includegraphics[width=\wid]{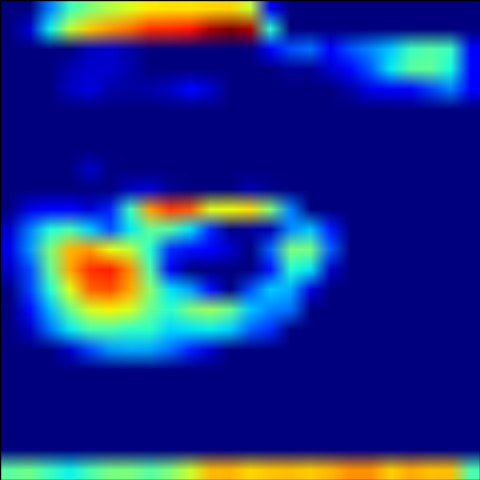}\quad
\includegraphics[width=\wid]{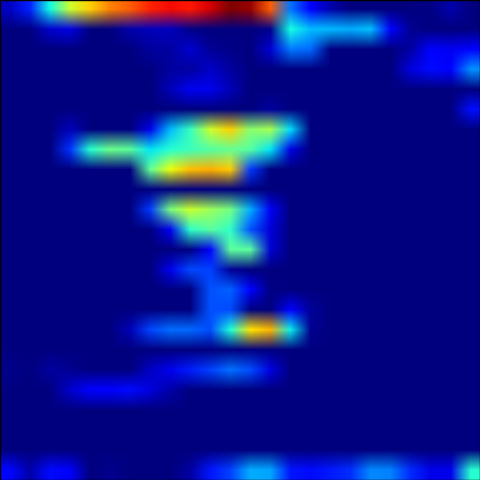}\quad
\includegraphics[width=\wid]{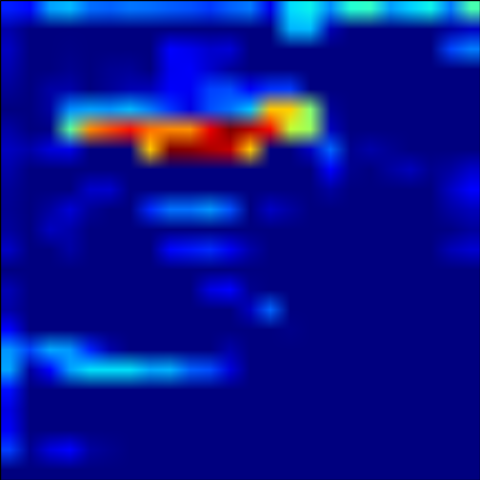}\quad
\includegraphics[width=\wid]{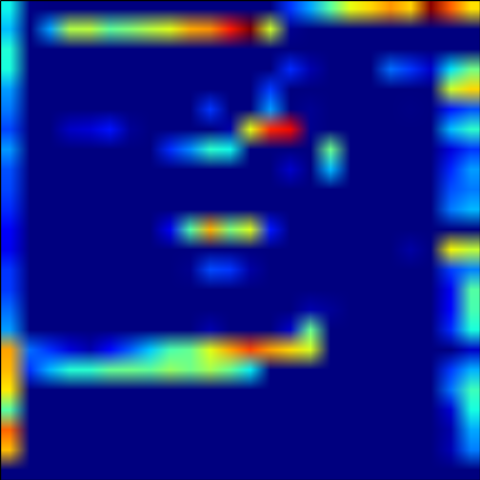}\quad
\includegraphics[width=\wid]{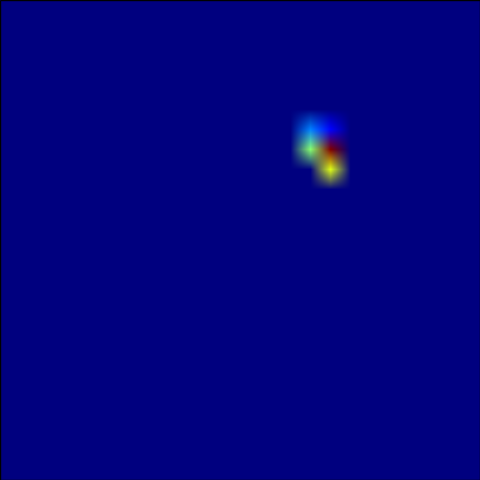}\\
\end{center}
\caption{Examples of feature maps of the first (top row) and
second (bottom row) convolutional layers computed from image
at the right of Figure \ref{fig:images}.}
\label{fig:featmaps}
\end{figure*}

The results of the experiments are summarized in Table \ref{tab:results}.
The first four rows describe the test set accuracies of the proposed
methods for the single-camera and the two-camera cases. One can
clearly see that both proposed approaches are relatively accurate in their
recognition. However, in both cases the DNN is superior to the shallow 
architecture, although the margin is not particularly large. In fact, the
results suggest that the car type recognizer uses relatively simple features
as a basis of detection. Also the feature maps of the DNN 
shown in Figure \ref{fig:featmaps} indicate
that the convolutional layers learn to highlight image patterns
that distinguish car types from each other: grille pattern, grille shape, 
headlight shape, etc. Possibly the added accuracy of the deep architecture
is due to the higher level features, such as the distance of headlights and so on.
Note that although the difference in accuracies may appear small, the
DNN in fact makes over 40\% fewer errors than the SVM in the two-camera case.
Moreover, a visual inspection of the erroneous classifications show
that the problems arise with vehicles "between two classes" difficult even
for humans to categorize, such as small vans, ambulances, etc.

The proposed methods are also compared to other approaches in Table \ref{tab:results}.
Although the database in each paper is separate, the numbers indicate that
the proposed method exceed the state of the art. In particular, the size of the
database in our case is larger than that of the other experiments thus
increasing our belief that data driven approaches are more reliable than
manually engineered feature extraction 
pipelines \cite{kafai2012dynamic,zhang2013vehicle,Lee-Neural-2013}. 

We do acknowledge that the comparison would be more appropriate
using the same database for all the methods. However, the datasets are 
unfortunately not public, and the implementation of the alternative (relatively
tailored) methods in exactly the original manner is a non-trivial 
task. However, as our database is among the largest, we have a strong
belief that the proposed data-driven approaches would be successful
with the other datasets, as well.
To facilitate later comparison in a reproducible manner,
we provide the details of our network (topology and pretrained
coefficients) at the supplementary site for the
paper\footnote{\url{http://www.cs.tut.fi/~hehu/CarType/}}.

\begin{figure*}
\begin{center}
\includegraphics[width=0.25\textwidth]{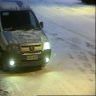}
\includegraphics[width=0.25\textwidth]{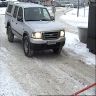}
\includegraphics[width=0.25\textwidth]{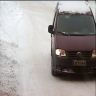}
\end{center}
\caption{Examples of vehicles incorrectly classified by the
network. All three are assigned to category \emph{van}, although
they are annotated as \emph{normal vehicle}.}
\label{fig:errors}
\end{figure*}

The erroneous recognition results are often related to ambiguous
annotation. Three examples of incorrectly classified results are
shown in Figure~\ref{fig:errors}. In all cases, the vehicles
are categorized as "normal vehicle" instead of the annotated 
category "van". In fact, they all represent a small van, which
has resemblance to both categories and are slightly difficult
to categorize unambiguously.

\section{Conclusions}
\label{sec:conclusions}

In this paper we studied the use of data driven image recognition 
techniques for car type classification. This separates the work from
existing approaches that rely on an elaborate manually engineered 
feature extraction pipeline, thus simplifying the software architecture
substantially. It was shown that both studied methods (Deep neural 
network and SVM with SIFT features) are able to accurately recognize
the car type, and the DNN is superior in accuracy to the SVM.

It was also shown that both proposed methods outperform methods of the
earlier studies. However, one should bear in mind that the
image databases in these studies were different and not directly
comparable with each other. In fact, one our our future plans is to
extend our work towards freely available image databases 
\cite{yang2015large,krause20133d}. However, this will require 
manual annotation of the databases, as the current annotations
only include the car make. 

An interesting question is how well the trained models generalize to
novel situations, such as new traffic lanes with slightly different
viewing angle, illumination or direction of traffic. In this paper
we have shown that a single model can learn to recognize vehicles 
on two lanes with highly varying illumination. Thus, there is no reason
why the network would not generalize to further installations. However,
the study of how many lanes need to be labeled manually in order to
generalize to any environment is left for future work on the topic.

\section{Acknowledgements}

The authors would like to acknowledge \emph{CSC - IT Center for Science Ltd.}
for computational resources.

\bibliographystyle{IEEEtran}
{
\bibliography{DNN}
}

\end{document}